\def\year{2018}\relax
\documentclass[letterpaper]{article} 
\usepackage{aaai18}  
\usepackage{times}  
\usepackage{helvet}  

\usepackage{courier}  
\usepackage{url}  
\usepackage{graphicx}  
\usepackage{url}
\usepackage{color}
\usepackage{float}
\usepackage{tabularx}
\usepackage{subfig}
\usepackage{multirow}
\usepackage{latexsym}
\usepackage{amsmath}
\usepackage{amssymb}
\usepackage{colortbl}
\usepackage{array}
\usepackage{bm}
\usepackage{booktabs}
\usepackage{pgfplots}
\usepackage{algorithm}
\usepackage{algorithmic}
\newcommand{\tabincell}[2]{\begin{tabular}{@{}#1@{}}#2\end{tabular}}

\newcommand{\citet}[1]{\citeauthor{#1}~(\citeyear{#1})}

\newcommand{\hytt}[1]{\texttt{\hyphenchar\font=\defaulthyphenchar #1}}

\pgfplotsset{compat=1.14}

\def\bb{\mathbf{b}}
\def\bc{\mathbf{c}}

\def\b1{\mathbf{1}}
\def\bbf{\textbf{f}}

\def\bh{\textbf{h}}

\def\softmax{\textbf{softmax}}

\def\bi{\mathbf{i}}

\def\bo{\mathbf{o}}

\def\bx{\mathbf{x}}
\def\by{\mathbf{y}}
\def\bz{\mathbf{z}}

\def\bW{\mathbf{W}}

\allowdisplaybreaks
\title{Meta Multi-Task Learning for Sequence Modeling}
\author{Junkun Chen, Xipeng Qiu\thanks{Corresponding Author}, Pengfei Liu, Xuanjing Huang\\
Shanghai Key Laboratory of Intelligent Information Processing, Fudan University\\
School of Computer Science, Fudan University\\
825 Zhangheng Road, Shanghai, China\\
\{jkchen16, xpqiu, pfliu14, xjhuang\}@fudan.edu.cn\\
}
\begin{document}
\maketitle
\begin{abstract}
Semantic composition functions have been playing a pivotal role in neural representation learning of text sequences. In spite of their success, most existing models suffer from the underfitting problem: they use the same shared compositional function on all the positions in the sequence, thereby lacking expressive power due to incapacity to capture the richness of compositionality. Besides, the composition functions of different tasks are independent and learned from scratch. In this paper, we propose a new sharing scheme of composition function across multiple tasks. Specifically, we use a shared meta-network to capture the meta-knowledge of semantic composition and generate the parameters of the task-specific semantic composition models.
We conduct extensive experiments on two types of tasks, text classification and sequence tagging, which demonstrate the benefits of our approach. Besides, we show that the shared meta-knowledge learned by our proposed model can be regarded as off-the-shelf knowledge and easily transferred to new tasks.
\end{abstract}

\section{Introduction}
Deep learning models have been widely used in many natural language processing (NLP) tasks. A major challenge is how to design and learn the semantic composition function while modeling a text sequence. The typical composition models involve sequential \cite{sutskever2014sequence,chung2014empirical}, convolutional \cite{collobert2011natural,kalchbrenner2014convolutional,kim2014convolutional} and syntactic \cite{socher2013recursive,tai2015improved,zhu2015long} compositional models.

In spite of their success, these models have two major limitations. First, they usually use a shared composition function for all kinds of semantic compositions, even though the compositions have different characteristics in nature. For example, the composition of the adjective and the noun differs significantly from the composition of the verb and the noun. Second, different composition functions are learned from scratch in different tasks. However, given a certain natural language, its composition functions should be the same (on meta-knowledge level at least), even if the tasks are different.

To address these problems, we need to design a dynamic composition function which can vary with different positions and contexts in a sequence, and share it across the different tasks.
To share some meta-knowledge of composition function, we can adopt the multi-task learning  \cite{caruana1997multitask}. However, the sharing scheme of most neural multi-task learning methods is \textit{feature-level sharing}, where a subspace of the feature space is shared across all the tasks.
Although these sharing schemes are successfully used in various NLP tasks \cite{collobert2008unified,luong2015multi,liu2015representation,liu2016recurrent,hashimoto2017joint,chen-EtAl:2017:Long2}, they are not suitable to share the composition function.

\begin{figure}[t]\centering
\subfloat[feature-level]{
  \includegraphics[width=0.45\linewidth]{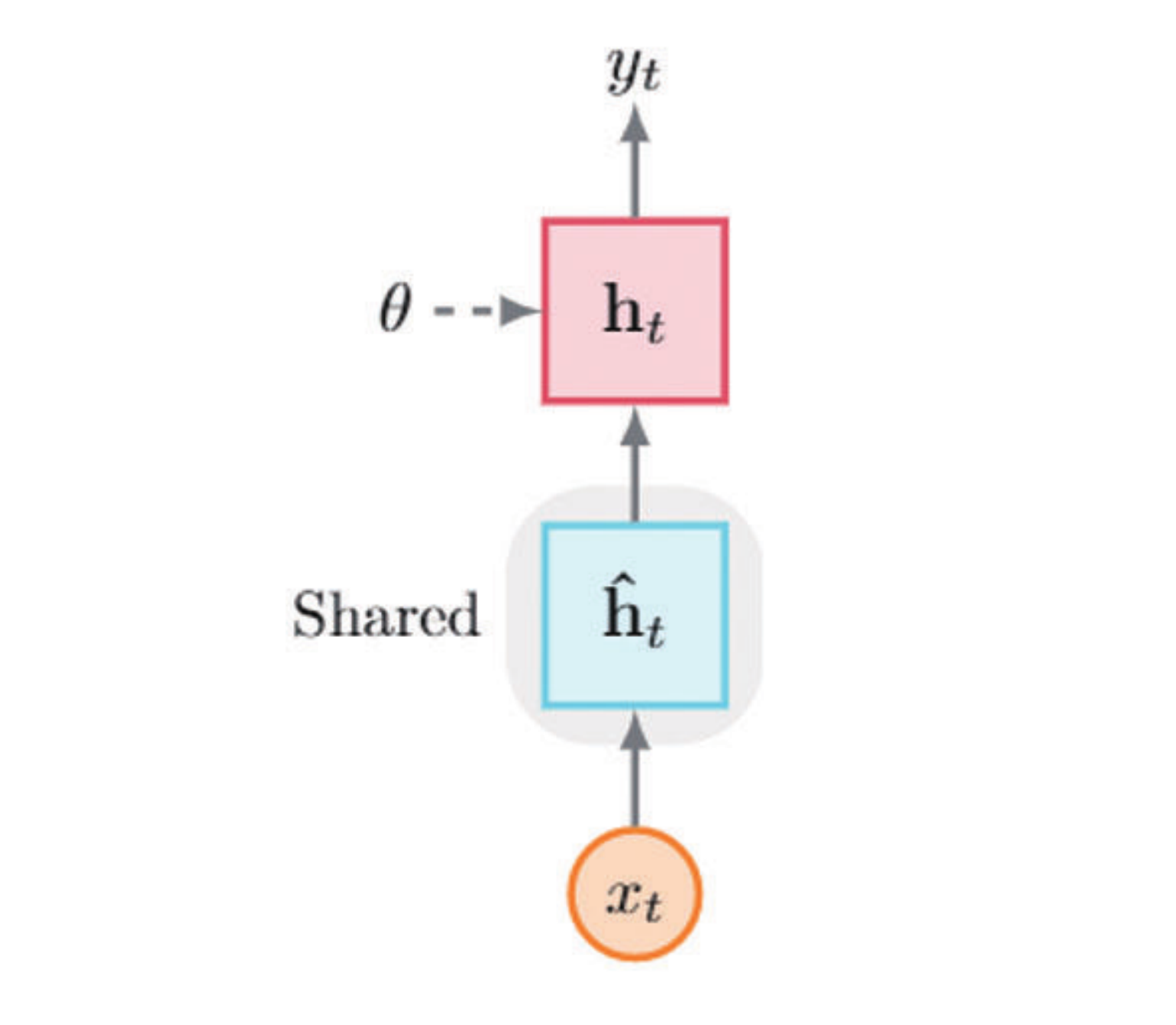}\label{fig:groups1}
  }
\subfloat[function-level]{
  \includegraphics[width=0.45\linewidth]{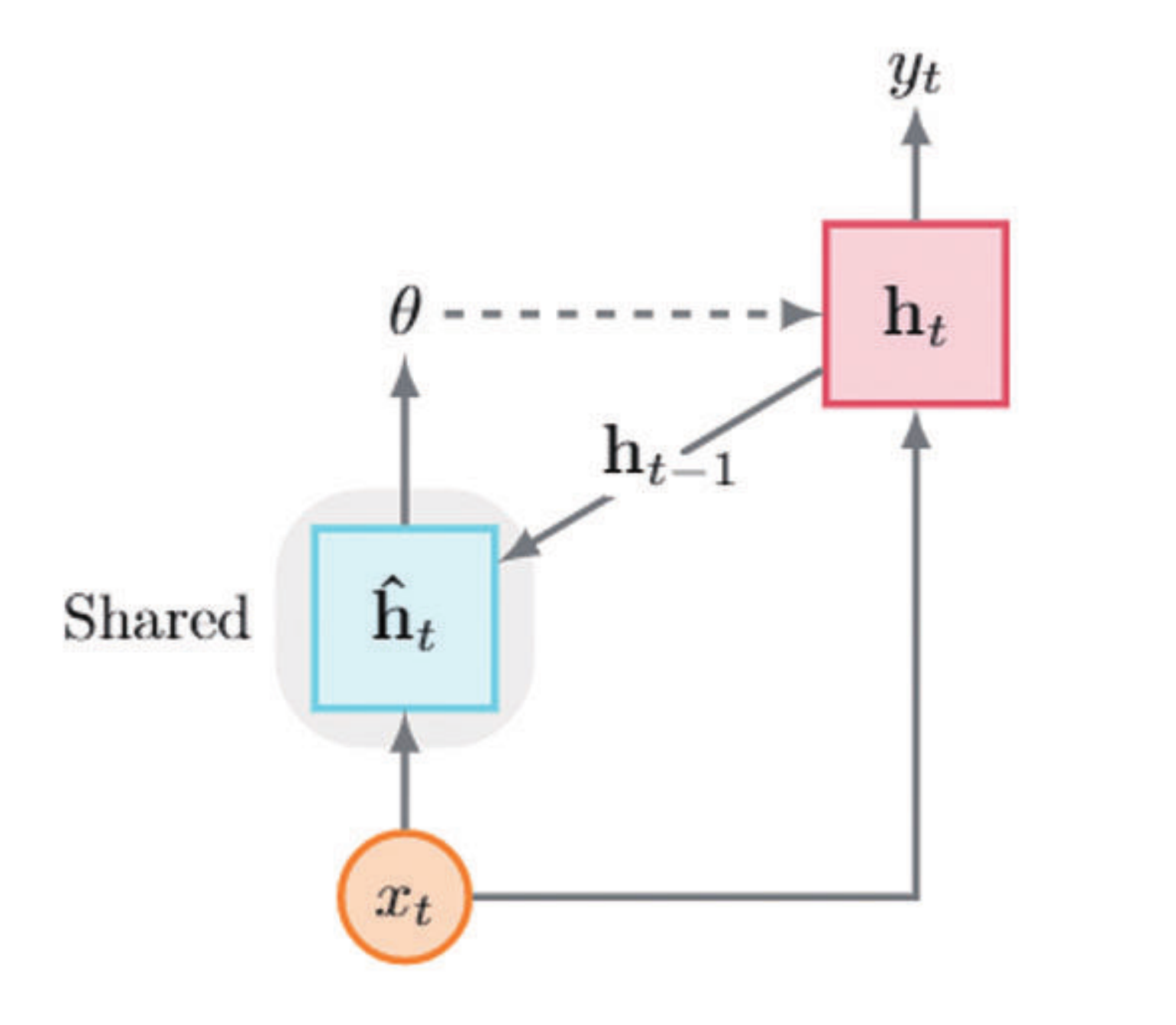}\label{fig:groups2}
  }
\caption{Two different sharing schemes. $\theta$ denotes the parameters of task-specific composition function. $x_t$, $y_t$, $h_t$ and $\hat{h}_t$ are the input, output, shared and private hidden states at step $t$. (a) The generic feature-level sharing scheme,  in which the shared features will be taken as inputs for task-specific layers. (b) Our proposed function-level sharing scheme, in which a shared Meta-LSTM controls the parameters $\theta_t$ of task-specific composition function.}\label{fig:share_scheme}
\end{figure}

In this paper, inspired by recent work on dynamic parameter generation \cite{de2016dynamic,bertinetto2016learning,ha2016hypernetworks}, we propose a \textit{function-level} sharing scheme for multi-task learning, in which a shared meta-network is used to learn the meta-knowledge of semantic composition among the different tasks. The task-specific semantic composition function is generated by the meta-network. Then the task-specific composition function is used to obtain the task-specific representation of a text sequence. The difference between two sharing schemes is shown in Figure \ref{fig:share_scheme}. Specifically, we use two LSTMs as meta and basic (task-specific) network respectively. The meta LSTM is shared for all the tasks. The parameters of the basic LSTM are generated based on the current context by the meta LSTM, therefore the composition function is not only task-specific but also position-specific. The whole network is differentiable with respect to the model parameters and can be trained end-to-end.

We demonstrate the effectiveness of our architectures on two kinds of NLP tasks: text classification and sequence tagging. Experimental results show that jointly learning of multiple related tasks can improve the performance of each task relative to learning them independently.

Our contributions are of three-folds:
\begin{itemize}
  \item We propose a new perspective of information sharing scheme for multi-task learning. Different from the feature-level sharing, we introduce \textit{function-level} sharing scheme to extract the meta knowledge of semantic composition across the different tasks.
  \item 
  The Meta-LSTMs not only improve the performance of multi-task learning, but also benefit the single-task learning since the parameters of the basic LSTM vary from position to position, in contrast to the same parameters used for all the positions in the standard LSTM. Thus, the of the task-specific LSTM vary from position to position, allowing for more sophisticated semantic compositions of text sequence.

  \item The Meta-LSTM can be regarded as a prior knowledge of semantic composition, while the basic LSTM is the posterior knowledge. Therefore, our learned Meta-LSTM  also provides an efficient way of performing transfer learning \cite{pan2010survey}. Under this view, a new task can no longer be simply seen as an isolated task that starts accumulating knowledge afresh. As more tasks are observed, the learning mechanism is expected to benefit from previous experience.
\end{itemize}

\section{Generic Neural Architecture of Multi-Task Learning for Sequence Modeling}
In this section, we briefly describe generic neural architecture of multi-task learning .

\subsection{Task Definition}

The task of Sequence Modeling is to assign a label sequence $Y=\{y_1,y_2,\cdots,y_T\}$. to a text sequence $X=\{x_1,x_2,\cdots,x_T\}$. In classification task, $Y$ is a single label.
Assuming that there are $K$ related tasks, we refer $\mathcal{D}_k$ as the corpus of the $k$-th task with $N_k$ samples:
\begin{equation}
\mathcal{D}_k = \{(X_i^{(k)},Y_i^{(k)})\}_{i=1}^{N_k},
\end{equation}
where $X_i^k$ and $Y_i^k$ denote the $i$-th sample and its label respectively in the $k$-th task.

Multi-task learning  \cite{caruana1997multitask} is an approach to learn multiple related tasks simultaneously to significantly improve performance relative to learning each task independently. The main challenge of multi-task learning is how to design the sharing scheme. For the shallow classifier with discrete features, it is relatively difficult to design the shared feature spaces, usually resulting in a complex model. Fortunately, deep neural models provide a convenient way to share information among multiple tasks.

\subsection{Generic Neural Architecture of Multi-Task Learning for Sequence Modeling}
\begin{figure}[t]
  \includegraphics[width=0.8\linewidth]{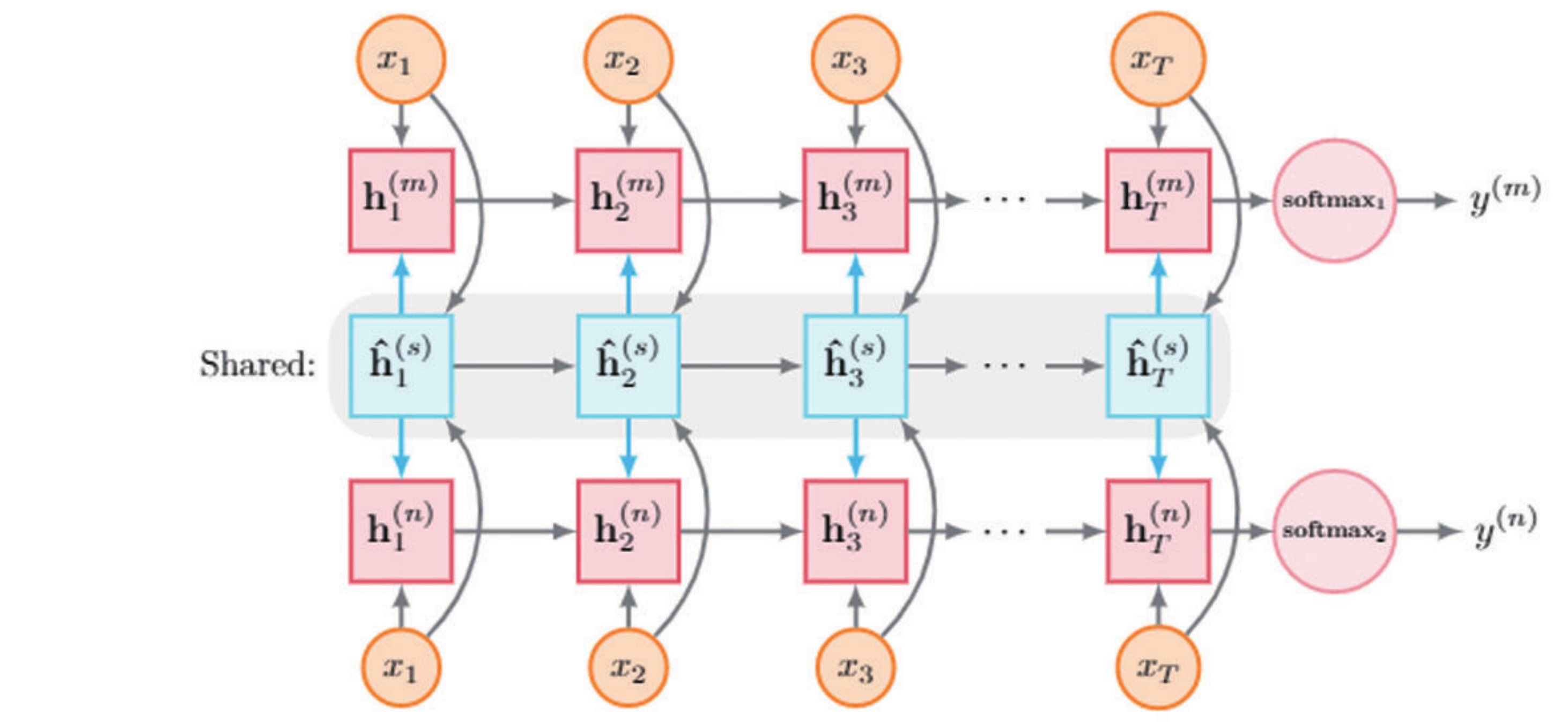}
  \caption{Generic Architecture of Multi-Task Learning. The blue modules whose output will be taken as the input of private layers are shared between different tasks.}\label{fig:generic}
\end{figure}

The generic neural architecture of multi-task learning is to share some lower layers to determine common features. After the shared layers, the remaining higher layers are parallel and independent respective to each specific task. Figure \ref{fig:generic} illustrates the  generic architecture of multi-task learning. \cite{collobert2008unified,liu2015representation,liu2016recurrent}

\subsubsection{Sequence Modeling with LSTM}
There are many neural sentence models, which can be used for sequence modeling, including recurrent neural networks \cite{sutskever2014sequence,chung2014empirical}, convolutional neural networks \cite{collobert2011natural,kalchbrenner2014convolutional}, and recursive neural networks \cite{socher2013recursive}.
Here we adopt recurrent neural network with long short-term memory (LSTM) due to their superior performance in various NLP tasks.

LSTM \cite{hochreiter1997long} is a type of recurrent neural network (RNN), and specifically addresses the issue of learning long-term dependencies.
While there are numerous LSTM variants, here we use the LSTM architecture used by \cite{jozefowicz2015empirical}, which is similar to the architecture of \cite{graves2013generating} but without peep-hole connections.

We define the LSTM \emph{units} at each time step $t$ to be a collection of vectors in $\mathbb{R}^h$: an \emph{input gate} $\bi_t$, a \emph{forget gate} $\bbf_t$,  an \emph{output gate} $\bo_t$, a \emph{memory cell} $\bc_t$ and a \emph{hidden state} $\bh_t$. $d$ is the number of the LSTM units. The elements of the gating vectors $\bi_t$, $\bbf_t$ and $\bo_t$ are in $[0, 1]$.

The LSTM is compactly specified as follows.

\begin{align}
	\begin{bmatrix}
		\mathbf{g}_{t} \\
		\mathbf{o}_{t} \\
		\mathbf{i}_{t} \\
		\mathbf{f}_{t}
	\end{bmatrix}
	&=
	\begin{bmatrix}
		\tanh \\
		\sigma \\
		\sigma \\
		\sigma
	\end{bmatrix}
    \begin{pmatrix}
	\bW
	\begin{bmatrix}
		\mathbf{x}_{t} \\
		\mathbf{h}_{t-1}
	\end{bmatrix}+\bb
    \end{pmatrix}, \label{eq:lstm1}\\
\mathbf{c}_{t} &=
		\mathbf{g}_{t} \odot \mathbf{i}_{t}
		+ \mathbf{c}_{t-1} \odot \mathbf{f}_{t}, \\
	\mathbf{h}_{t} &= \mathbf{o}_{t}  \odot \tanh\left( \mathbf{c}_{t}  \right)\label{eq:lstm3},
\end{align}
where $\bx_t \in \mathbb{R}^{d}$ is the input at the current time step;
$\bW \in \mathbb{R}^{4h\times(h+d)}$ and $\bb \in \mathbb{R}^{4h}$ are parameters of affine transformation;
$\sigma$ denotes the logistic sigmoid function and $\odot$ denotes elementwise multiplication.

The update of each LSTM unit can be written precisely as follows:
\begin{align}
\bh_t &= \mathbf{LSTM}(\bh_{t-1},\mathbf{x}_t, \theta).  \label{eq:LSTM}
\end{align}
Here, the function $\mathbf{LSTM}(\cdot, \cdot, \cdot)$ is a shorthand for Eq. (\ref{eq:lstm1}-\ref{eq:lstm3}), and $\theta$ represents all the parameters of LSTM.

Given a text sequence $X = \{x_1, x_2, \cdots, x_T\}$, we first use a lookup layer to get the vector representation (embeddings) $\bx_t$ of each word $x_t$.
The output at the last moment $\bh_T$  can be regarded as the representation of the whole sequence.


\subsubsection{Shared-Private Sharing Scheme}

To exploit the shared information between these different tasks, the general deep multi-task architecture consists of a private (task-specific) layer and a shared (task-invariant) layer.
The shared layer captures the shared information for all the tasks.

The shared layer and private layer is arranged in stacked manner. The private layer takes the output of the shared layer as input. For task $k$, the hidden states of shared layer and private layer are:
\begin{align}
\bh^{(s)}_t& = \text{LSTM}(\bx_{t}, \bh^{(s)}_{t-1},\theta_s),\label{eq:m2-1}\\
\bh^{(k)}_t &= \text{LSTM}(\begin{bmatrix}
                                      \bx_{t}\\
                                      \bh^{(s)}_t
                                      \end{bmatrix}, \bh^{(k)}_{t-1},\theta_k)\label{eq:m2-2}
\end{align}
where $\bh^{(s)}_t$ and $\bh^{(k)}_t$ are hidden states of the shared layer and the $k$-th task-specific layer respectively; $\theta_s$ and $\theta_k$ denote their parameters.

\subsubsection{Task-specific Output Layer}
The task-specific representations $\bh^{(k)}$, which is emitted by the multi-task architecture, are ultimately fed into different task-specific output layers. 

Here, we use two kinds of tasks: text classification and sequence tagging.
\paragraph{Text Classification}

For task $k$ in , the label predictor is defined as
\begin{align}
{\hat{\by}}^{(k)} = \softmax(\bW^{(k)}\bh^{(k)} + \bb^{(k)}),
\end{align}
where ${\hat{\by}}^{(k)}$ is prediction probabilities for task $k$, $\bW^{(k)}$ is the weight matrix which needs to be learned, and $\bb^{(k)}$ is a bias term.

\paragraph{Sequence Tagging}
Following the idea of \cite{huang2015bidirectional,ma2016end}, we use a conditional random field (CRF) \cite{lafferty2001conditional} as output layer.

\subsection{Training}

The parameters of the network are trained to minimise the cross-entropy of the predicted and true distributions for all tasks.
\begin{align}
\mathcal{L}(\Theta) = -\sum_{k=1}^{K} {\lambda}_k  \sum_{i=1}^{N_k} \by_i^{(k)} \log(\hat{\by}_i^{(k)}),
\end{align}
where $\lambda_k$ is the weights for each task $k$ respectively; $\by_i^{(k)}$ is the one-hot vector of the ground-truth label of the sample $X_i^{(k)}$; $\hat{y}_i^{(k)}$ is its prediction probabilities.

It is worth noticing that labeled data for training each task can come from completely different datasets. Following \cite{collobert2008unified}, the training is achieved in a stochastic manner by looping over the tasks:
\begin{enumerate}
  \item Select a random task.
  \item Select a mini-batch of examples from this task.
  \item Update the parameters for this task by taking a gradient step with respect to this mini-batch.
  \item Go to 1.
\end{enumerate}

After the joint learning phase, we can use a fine tuning strategy to further optimize the performance for each task.

\section{Meta Multi-Task Learning}

In this paper, we take a very different multi-task architecture from meta-learning perspective \cite{brazdil2008metalearning}. One goal of meta-learning is to find efficient mechanisms to transfer knowledge across domains or tasks \cite{lemke2015metalearning}.

\begin{figure}[t]\centering
 \includegraphics[width=0.98\linewidth]{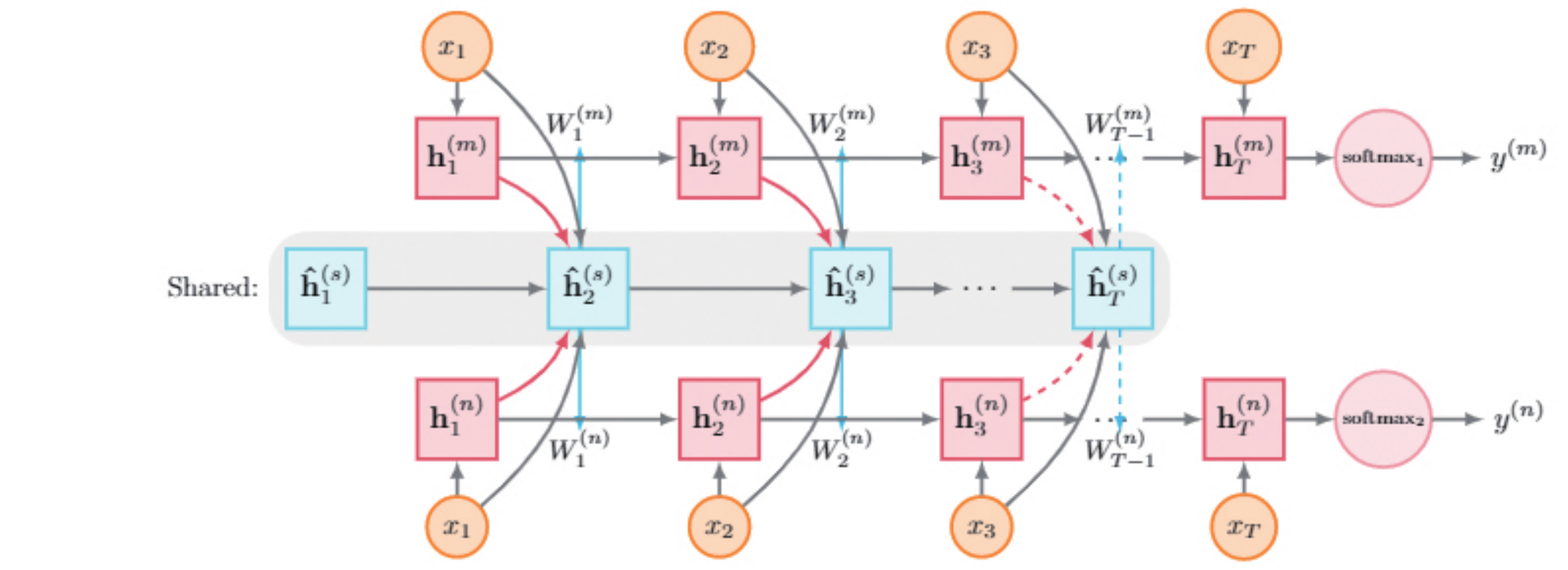}
 \caption{Architecture of Meta Multi-task Learning. The blue modules are shared between different tasks, which control the parameters of private layers.
 }\label{fig:match}
\end{figure}

Different from the generic architecture with the \textit{representational sharing} (feature sharing) scheme, our proposed architecture uses a \textit{functional sharing} scheme, which consists of two kinds of networks. As shown in Figure \ref{fig:match}, for each task, a basic network is used for task-specific prediction, whose parameters are controlled by a shared meta network across all the tasks.

We firstly introduce our architecture on single task, then apply it for multi-task learning.

\subsection{Meta-LSTMs for Single Task}
Inspired by recent work on dynamic parameter prediction \cite{de2016dynamic,bertinetto2016learning,ha2016hypernetworks}, we also use a meta network to generate the parameters of the task network (basic network). Specific to text classification, we use LSTM for both the networks in this paper, but other options are possible.

There are two networks for each single task: a basic LSTM and a meta LSTM.

\paragraph{Basic-LSTM}
For each specific task, we use a basic LSTM to encode the text sequence. Different from the standard LSTM, the parameters of the basic LSTM is controlled by a meta vector $\bz_t$, generated by the meta LSTM.
The new equations of the basic LSTM are
\begin{align}
	\begin{bmatrix}
		\mathbf{g}_{t} \\
		\mathbf{o}_{t} \\
		\mathbf{i}_{t} \\
		\mathbf{f}_{t}
	\end{bmatrix}
	&=
	\begin{bmatrix}
		\tanh \\
		\sigma \\
		\sigma \\
		\sigma
	\end{bmatrix}
    \begin{pmatrix}
	\bW(\bz_t)
	\begin{bmatrix}
		\mathbf{x}_{t} \\
		\mathbf{h}_{t-1}
	\end{bmatrix}
+ \bb(\bz_t)
    \end{pmatrix}, \label{eq:lstm5}\\
\mathbf{c}_{t} &=
		\mathbf{g}_{t} \odot \mathbf{i}_{t}
		+ \mathbf{c}_{t-1} \odot \mathbf{f}_{t}, \\
	\mathbf{h}_{t} &= \mathbf{o}_{t}  \odot \tanh\left( \mathbf{c}_{t}  \right)\label{eq:lstm6},
\end{align}
where $\bW(\bz_t): \mathbb{R}^z\rightarrow \mathbb{R}^{4h\times (h+d)}$ and $\bb(\bz_t): \mathbb{R}^z\rightarrow \mathbb{R}^{4h}$ are dynamic parameters controlled by the meta network.

Since the output space of the dynamic parameters $\bW(\bz_t)$ is very large, its computation is slow without considering matrix optimization algorithms. Moreover, the large parameters makes the model suffer from the risk of overfitting. To remedy this, we define $\bW(\bz_t)$ with a low-rank factorized representation of the weights, analogous to the Singular Value Decomposition.

The parameters $\bW(\bz_t)$ and $\bb(\bz_t)$ of the basic LSTM are computed by
\begin{align}
 \mathbf{W}(\bz_t) &= \begin{bmatrix}
    P_c \mathbf{D}(\bz_t) Q_c\\
    P_o \mathbf{D}(\bz_t) Q_o\\
    P_i \mathbf{D}(\bz_t) Q_i\\
    P_f \mathbf{D}(\bz_t) Q_f
    \label{eq:get_matrix}
    \end{bmatrix}\\
\bb(\bz_t)&=\begin{bmatrix}
	B_c \bz_t\\
    B_o \bz_t\\
    B_i \bz_t\\
    B_f \bz_t
	\end{bmatrix}
\end{align}
where $P_*\in \mathbb{R}^{h\times z}$, $Q_*\in \mathbb{R}^{z\times d}$ and $B_*\in \mathbb{R}^{h\times z}$ are parameters for $*\in \{c,o,i,f\}$.

Thus, our basic LSTM needs $(8hz + 4dz)$ parameters, while the standard LSTM has $(4h^2+4hd+4h)$ parameters.
With a small $z$, the basic LSTM needs less parameters than the standard LSTM. For example, if we set $d = h = 100$ and $z=20$, our basic LSTM just needs $24,000$ parameter while the standard LSTM needs $80,400$ parameters.

\paragraph{Meta-LSTM}

The Meta-LSTM is usually a smaller network, which depends on the input $\bx_t$ and the previous hidden state $\bh_{t-1}$ of the basic LSTM.

The Meta-LSTM cell is given by:
\begin{align}
	\begin{bmatrix}
		\hat{\mathbf{g}}_{t} \\
		\hat{\mathbf{o}}_{t} \\
		\hat{\mathbf{i}}_{t} \\
		\hat{\mathbf{f}}_{t}
	\end{bmatrix}
	&=
	\begin{bmatrix}
		\tanh \\
		\sigma \\
		\sigma \\
		\sigma
	\end{bmatrix}
    \begin{pmatrix}
	\bW_m
	\begin{bmatrix}
		\mathbf{x}_{t} \\
        \mathbf{\hat{h}}_{t-1}\\
		\mathbf{h}_{t-1}
	\end{bmatrix}+\bb_m
    \end{pmatrix}, \label{eq:lstm31}\\
\hat{\mathbf{c}}_{t} &=
		\hat{\mathbf{g}}_{t} \odot \hat{\mathbf{i}}_{t}
		+ \hat{\mathbf{c}}_{t-1} \odot \hat{\mathbf{f}}_{t}, \\
	\hat{\mathbf{h}}_{t} &= \hat{\mathbf{o}}_{t}  \odot \tanh\left( \hat{\mathbf{c}}_{t}  \right)\label{eq:lstm33},\\
\bz_t &= \mathbf{W}_z \hat{\mathbf{h}}_{t},
\end{align}
where $\mathbf{W}_{m} \in \mathbb{R}^{4m \times (d+h+m)}$ and $\mathbf{b}_{m} \in \mathbb{R}^{4m}$ are parameters of Meta-LSTM; $\mathbf{W}_{z} \in \mathbb{R}^{z \times m}$ is a transformation matrix.

Thus, the Meta-LSTM needs $(4m(d+h+m+1)+mz)$ parameters. When $d = h = 100$ and $z=m=20$, its parameter number is $18,080$. The total parameter number of the whole networks is $42,080$, nearly half of the standard LSTM.

We precisely describe the update of the units of the Meta-LSTMs  as follows:
\begin{align}
[\hat{\bh}_t , \bz_t] & = \text{Meta-LSTM}(\bx_{t}, \hat{\bh}_{t-1},\bh_{t-1};\theta_m),\\
\bh_t &= \text{Basic-LSTM}(\bx_{t}, \bh_{t-1};\bz_t, \theta_b)
\end{align}
where $\theta_m$ and $\theta_b$ denote the parameters of the Meta-LSTM and Basic-LSTM respectively.

Compared to the standard LSTM, the Meta-LSTMs have two advantages. One is the parameters of the Basic-LSTM is dynamically generated conditioned on the input at the position, while the parameters of the standard LSTM are the same for all the positions, even though different positions have very different characteristics. Another is that the Meta-LSTMs usually have less parameters than the standard LSTM.


\subsection{Meta-LSTMs for Multi-Task Learning}

For multi-task learning, we can assign a basic network to each task, while sharing a meta network among tasks. The meta network captures the meta (shared) knowledge of different tasks.
The meta network can learn at the “meta-level” of predicting parameters for the basic task-specific network.

For task $k$, the hidden states of the shared layer and the private layer are:
\begin{align}
[\hat{\bh}^{(s)}_t, \bz^{(s)}_t]& = \text{Meta-LSTM}(\bx_{t}, \hat{\bh}^{(s)}_{t-1},\bh^{(k)}_{t-1};\theta^{(s)}_m),\label{eq:m12-1}\\
\bh^{(k)}_t &= \text{Basic-LSTM}(\bx_{t}, \bh^{(k)}_{t-1};\bz^{(s)}_t, \theta^{(k)}_b)\label{eq:m12-2}
\end{align}
where $\hat{\bh}^{(s)}_t$ and $\bh^{(k)}_t$ are the hidden states of the shared meta LSTM and the $k$-th task-specific basic LSTM respectively; $\theta^{(s)}_m$ and $\theta^{(k)}_b$ denote their parameters. The superscript $(s)$ indicates the parameters or variables are shared across the different tasks.

\section{Experiment}

In this section, we investigate the empirical performances of our proposed model on two multi-task datasets. Each dataset contains several related tasks.

\subsection{Exp-I: Multi-task Learning of text classification}

We first conduct our experiment on classification tasks.

\paragraph{Datasets}
\begin{table}[ht]\setlength{\tabcolsep}{3pt} \footnotesize
\centering
\begin{tabular}{|c|c|c|c|c|c|c|}
\hline
Datasets  & \tabincell{c}{Train\\Size} & \tabincell{c}{Dev.\\Size} & \tabincell{c}{Test\\Size} & Class & \tabincell{c}{Avg.\\Length} & \tabincell{c}{Voc.\\Size}  \\
\hline
Books           &   1400  & 200 & 400 & 2 & 159 & 62K\\
Elec            &   1398  & 200 & 400 & 2 & 101 & 30K\\
DVD             &	1400  & 200 & 400 & 2 & 173 & 69K\\
Kitchen         &	1400  & 200 & 400 & 2 & 89  & 28K\\
Apparel         &	1400  & 200 & 400 &	2 & 57  & 21K\\
Camera          &	1397  & 200 & 400 & 2 & 130 & 26K\\
Health          &	1400  &	200 & 400 & 2 & 81  & 26K\\
Music           &	1400  &	200 & 400 & 2 & 136 & 60K\\
Toys            &	1400  &	200 & 400 & 2 & 90  & 28K\\
Video           &	1400  &	200 & 400 & 2 & 156 & 57K\\
Baby            &	1300  &	200 & 400 & 2 & 104 & 26K\\
Mag             &	1370  &	200 & 400 & 2 & 117 & 30K\\
Soft           &	1315  &	200 & 400 & 2  & 129 & 26K\\
Sports          &	1400  &	200 & 400 & 2 & 94  & 30K\\
\hline
IMDB            &	1400  &	200 & 400 & 2 & 269 & 44K\\
MR              &	1400  &	200 & 400 & 2 & 21  & 12K\\
\hline


\end{tabular}
\caption{Statistics of sixteen multi-task datasets for text classification.} \label{tab:data}
\end{table}

For classification task, we test our model on 16 classification datasets, the first 14 datasets are product reviews that collected based on the dataset\footnote{\url{https://www.cs.jhu.edu/~mdredze/datasets/sentiment/}}, constructed by \citet{blitzer2007biographies}, contains Amazon product reviews from different domains: Books, DVDs, Electronics and Kitchen and so on. The goal in each domain is to classify a product review as either positive or negative. The datasets in each domain are partitioned randomly into training data, development data and testing data with the proportion of 70\%, 10\% and 20\% respectively. The detailed statistics are listed in Table \ref{tab:data}.

The remaining two datasets are two sub-datasets about movie reviews.
\begin{itemize}
  \item \textbf{IMDB} The movie reviews\footnote{\url{https://www.cs.jhu.edu/~mdredze/datasets/sentiment/unprocessed.tar.gz}} with labels of subjective or objective \cite{maas2011learning}.
  \item \textbf{MR} The movie reviews\footnote{\url{https://www.cs.cornell.edu/people/pabo/movie-review-data/}.} with two classes \cite{pang2005seeing}.
\end{itemize}

\paragraph{Competitor Models}
For single-task learning, we compare our Meta-LSTMs with three models.
\begin{itemize}
  \item \textbf{LSTM}: the standard LSTM with one hidden layer;
  \item \textbf{HyperLSTMs}: a similar model which also uses a small network to generate the weights for a larger network \cite{ha2016hypernetworks}.
\end{itemize}
For multi-task learning, we compare our Meta-LSTMs with the generic shared-private sharing scheme.
\begin{itemize}

  \item \textbf{ASP-MTL}: Proposed by \cite{liu2017adversarial}, using adversarial training method on PSP-MTL. 
  \item \textbf{PSP-MTL}: Parallel shared-private sharing scheme, using a fully-shared LSTM to extract features for all tasks and concatenate with the outputs from task-specific LSTM.
  \item \textbf{SSP-MTL}: Stacked shared-private sharing scheme, introduced in Section 2.\
\end{itemize}

\paragraph{Hyperparameters and Training}

\begin{table}[t]  \setlength{\tabcolsep}{3pt} 
\centering
\begin{tabular}{|l|*{1}{p{0.25\linewidth}|}}
    \hline
   Hyper-parameters & classification \\\hline
    Embedding dimension: $d$  & 200 \\
    Size of  $\bh$ in Basic-LSTM: $h$ & 100 \\
    Size of $\hat{\bh}$ in Meta-LSTM: $m$  & 40 \\
    Size of  meta vector $\bz$: $z$  & 40 \\
    Initial learning rate & 0.1 \\
    Regularization & $1E{-5}$ \\
    \hline
\end{tabular}
\caption{Hyper-parameters of our models.}\label{tab:paramSet}
\end{table}

The networks are trained with backpropagation and the gradient-based optimization is performed using the Adagrad update rule \cite{duchi2011adaptive}.

The word embeddings for all of the models are initialized with the 200d GloVe vectors (6B token version, \cite{pennington2014glove}) and fine-tuned during training to improve the performance.
The mini-batch size is set to 16.
The final hyper-parameters are set as Table \ref{tab:paramSet}.


\begin{table*}[t]\setlength{\tabcolsep}{3pt} \small
\center
\begin{tabular}{l*{9}{c}}
\toprule
\multirow{2}{*}{\textbf{Task}}  & \multicolumn{4}{c}{\textbf{Single Task}} & \multicolumn{4}{c}{\textbf{Multiple Tasks}} &   \multicolumn{1}{c}{\textbf{Transfer}} \\
\cmidrule(lr){2-5}  \cmidrule(lr){6-9} \cmidrule(lr){10-10}
 & LSTM & HyperLSTM & MetaLSTM & Avg.  & ASP-MTL$^*$ & PSP-MTL & SSP-MTL & Meta-MTL(ours) & Meta-MTL(ours)\\
\midrule
\multicolumn{1}{l}{Books}           & 79.5 & 78.3 & 83.0 & 80.2 & 87.0 & 84.3 & 85.3  & 87.5 & 86.3\\
\multicolumn{1}{l}{Electronics}     & 80.5 & 80.7 & 82.3 & 81.2 & 89.0 & 85.7 & 87.5  & 89.5 & 86.0\\
\multicolumn{1}{l}{DVD}             & 81.7 & 80.3 & 82.3 & 81.4 & 87.4 & 83.0 & 86.5  & 88.0 & 86.5\\
\multicolumn{1}{l}{Kitchen}         & 78.0 & 80.0 & 83.3 & 80.4 & 87.2 & 84.5 & 86.5  & 91.3 & 86.3\\
\multicolumn{1}{l}{Apparel}         & 83.2 & 85.8 & 86.5 & 85.2 & 88.7 & 83.7 & 86.0  & 87.0 & 86.0\\
\multicolumn{1}{l}{Camera}          & 85.2 & 88.3 & 88.3 & 87.2 & 91.3 & 86.5 & 87.5  & 89.7 & 87.0\\
\multicolumn{1}{l}{Health}          & 84.5 & 84.0 & 86.3 & 84.9 & 88.1 & 86.5 & 87.5  & 90.3 & 88.7\\
\multicolumn{1}{l}{Music}           & 76.7 & 78.5 & 80.0 & 78.4 & 82.6 & 81.3 & 85.7  & 86.3 & 85.7\\
\multicolumn{1}{l}{Toys}            & 83.2 & 83.7 & 84.3 & 83.7 & 88.8 & 83.5 & 87.0  & 88.5 & 85.3\\
\multicolumn{1}{l}{Video}           & 81.5 & 83.7 & 84.3 & 83.1 & 85.5 & 83.3 & 85.5  & 88.3 & 85.5\\
\multicolumn{1}{l}{Baby}            & 84.7 & 85.5 & 84.0 & 84.7 & 89.8 & 86.5 & 87.0  & 88.0 & 86.0\\
\multicolumn{1}{l}{Magazines}       & 89.2 & 91.3 & 92.3 & 90.9 & 92.4 & 88.3 & 88.0  & 91.0 & 90.3\\
\multicolumn{1}{l}{Software}        & 84.7 & 86.5 & 88.3 & 86.5 & 87.3 & 84.0 & 86.0  & 88.5 & 86.5\\
\multicolumn{1}{l}{Sports}          & 81.7 & 82.0 & 82.5 & 82.1 & 86.7 & 82.0 & 85.0  & 86.7 & 85.7\\
\multicolumn{1}{l}{IMDB}            & 81.7 & 77.0 & 83.5 & 80.7  & 85.8 & 82.0 & 84.5 & 88.0 & 87.3\\
\multicolumn{1}{l}{MR}              & 72.7 & 73.0 & 74.3 & 73.3 & 77.3 & 74.5 & 75.7  & 77.0 & 75.5\\
\midrule
\rowcolor[gray]{.9}
\multicolumn{1}{l}{AVG}             & 81.8 & 82.4 & 84.0 & 82.8  & 87.2$_{(+4.4)}$ & 83.7$_{(+0.9)}$ & 85.7$_{(+2.9)}$ & \textbf{87.9}$_{(+5.1)}$ &  85.9$_{(+3.1)}$\\\hline
\multicolumn{1}{l}{Parameters}    & 120$K$ & 321$K$ & 134$K$ &  & 5490$k$ & 2056$K$ & 1411$K$  & 1339$K$ & 1339$K$\\
\bottomrule
\end{tabular}
\caption{
Accuracies of our models on 16 datasets against typical baselines. The numbers in brackets represent the improvements relative to the average performance (Avg.) of three single task baselines. $^*$is from \cite{liu2017adversarial}
}\label{tab:mtl-16task}
\end{table*}

\begin{figure}[t]
  \centering
  \includegraphics[width=0.47\textwidth]{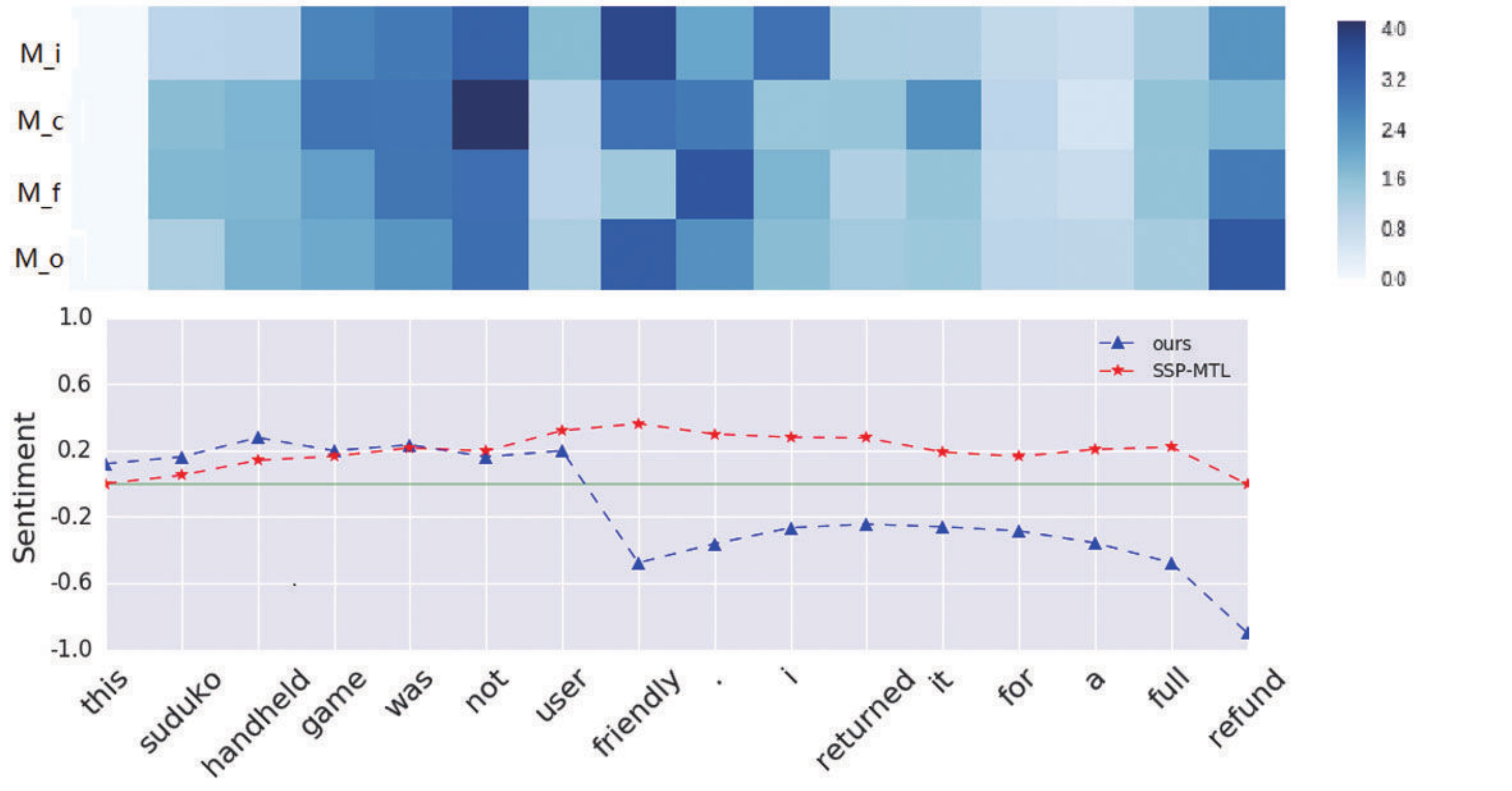}

  \caption{The lower figure presents sentiment prediction for each time step. Y-axis represents the predicted score, greater than zero for positive, less than zero for negative. The other one presents changes of the matrix generated by Eq.(\ref{eq:get_matrix}). We split the matrix into four pieces, which represent the matrix to compute input gate, new input, forget gate, and output gate respectively. And we calculate the changes of matrices each time step.}\label{fig:visual}
\end{figure}

\paragraph{Experiment result}
Table \ref{tab:mtl-16task} shows the classification accuracies on the tasks of product reviews.

The row of ``Single Task'' shows the results for single-task learning. With the help of Meta-LSTMs, the performances of the 16 subtasks are improved by an average of $3.2\%$, compared to the standard LSTM. However, the number of parameters is a little more than standard LSTM and much less than the HyperLSTMs.

For multi-task Learning, our model also achieves a better performance than our competitor models, with an average improvement of $5.1\%$ to average accuracy of single task and $2.2\%$ to best competitor Multi-task model. The main reason is that our models can capture more abstractive shared information. With a meta LSTM to generate the matrices, the layer will become more flexible.

With the meta network, our model can use quite a few parameters to achieve the state-of-the-art performances.

We have experimented various $z$ size in our multi-task model, where $z \in [20, 30,...,60]$, and the difference of the average accuracies of sixteen datasets is less than \textbf{$0.8\%$},  which indicates that the meta network with less parameters can also generate a basic network with a considerable good performance.

\paragraph{Visualization}
To illustrates the insight of our model, we randomly sample a sequence from the development set of Toys task. In Figure \ref{fig:visual} we predict the sentiment scores each time step. Moreover, to describe how our model works, we visualize the changes of matrices generated by Meta-LSTM, the changes $\textbf{diff}$ are calculate by Eq.\ref{eq:get_change}.

As we see it, the matrices change obviously facing the emotional vocabulary like ''\hytt{friendly}'', ''\hytt{refund}'', and slowly change to a normal state. They can also capture words that affect sentiments like ''\hytt{not}''. For this case, SSP-MTL give a wrong answer, it captures the emotion word''\hytt{refund}'', but it makes an error on pattern ''\hytt{not user friendly}'', we consider that it's because fixed matrices don't have satisfactorily ability to capture long patterns' emotions and information. Dynamic matrices generated by Meta-LSTM will make the layer more flexible.
\begin{align}
\mathbf{diff}^{(k)} = \textbf{mean}(\frac{\textbf{abs}(\bW^{(k)}-\bW^{(k-1)})}{\textbf{abs}(\bW^{(k-1)})}),
\label{eq:get_change}
\end{align}



\begin{figure}[t]\centering
  \includegraphics[width=0.8\linewidth]{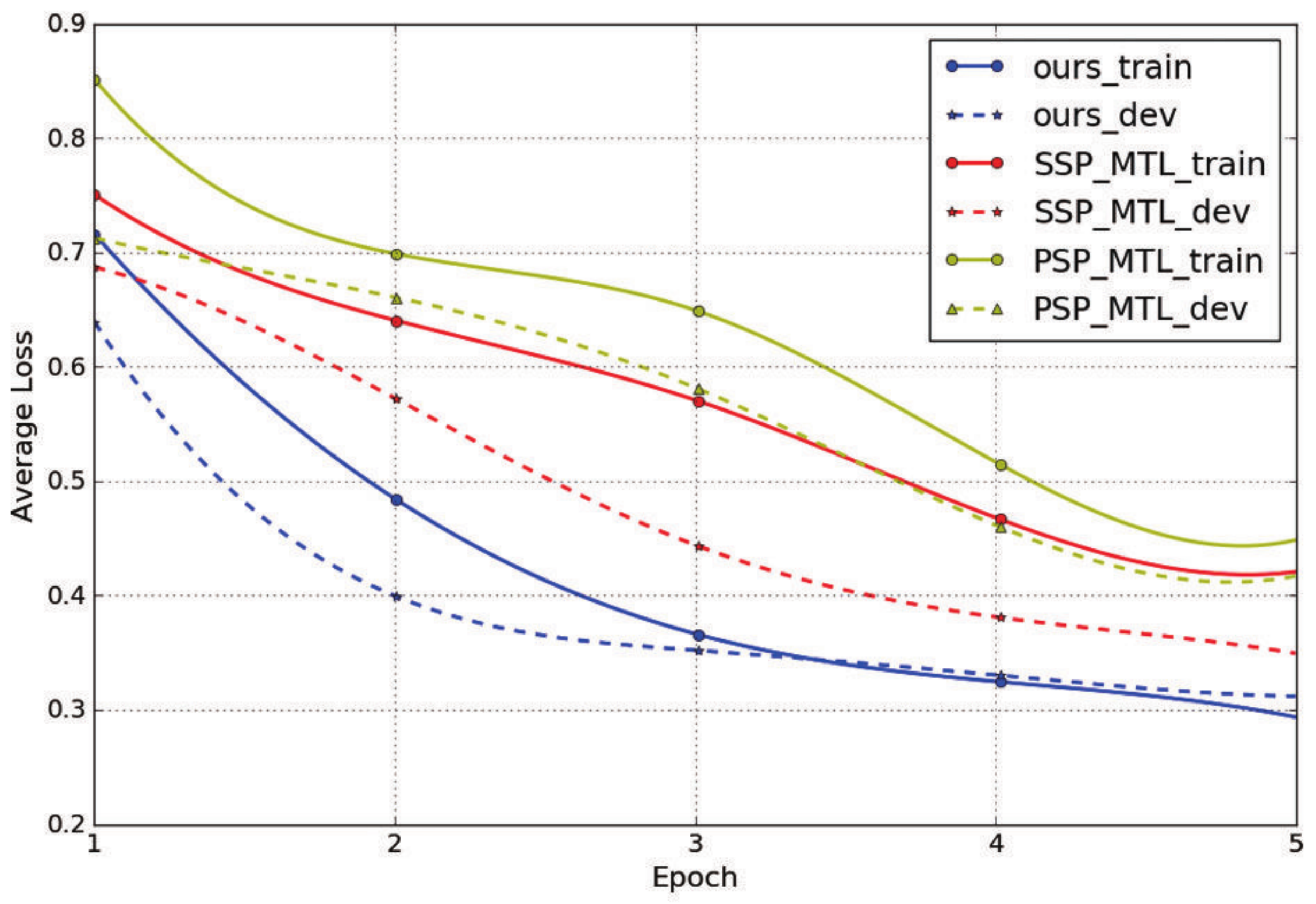}
  \caption{The train loss and dev loss of various multitask model decaying during the share training epochs.}\label{fig:loss}
\end{figure}

\paragraph{Convergence speed during shared training}
Figure \ref{fig:loss} shows the learning curves of various multi-task model on the 16 classification datasets.

Because it's inappropriate to evaluate different tasks every training step during shared parameters training since mini-batch of which tasks are selected randomly, so we use the average loss after every epoch. We can find that our proposed model is more efficient to fit the train datasets than our competitor models, and get better performance on the dev datasets. Therefore, we can consider that our model could learn shareable knowledge more effectively.

\paragraph{Meta knowledge transfer}
Since our Meta-LSTM captures some meta knowledge of semantic composition, which should have an ability of being transfered to a new task.
Under this view, a new task can no longer be simply seen as an isolated task that starts accumulating knowledge afresh. As more tasks are observed, the learning mechanism is expected to benefit from previous experience.

The meta network can be considered as off-the-shelf knowledge and then be used for unseen new tasks.

To test the transferability of our learned Meta-LSTM, we also design an experiment, in which we take turns choosing $15$ tasks to train our model with multi-task learning, then the learned Meta-LSTM are transferred to the remaining one task. The parameters of transferred Meta-LSTM, $\theta^{(s)}_m$ in Eq.(\ref{eq:m12-1}), are fixed and cannot be updated on the new task.

The results are also shown in the last column of Table \ref{tab:mtl-16task}.
With the help of meta knowledge, we observe an average improvement of $3.1\%$ over the average accuracy of single models, and even better than other competitor multi-task models. This observation indicates that we can save the meta knowledge into a meta network, which is quite useful for a new task.

\begin{table}[t]\setlength{\tabcolsep}{3pt} \small\footnotesize
\centering
\begin{tabular}{|*{5}{c|}}
\hline
\textbf{Tagging Dataset}  & \textbf{Task} & \textbf{Training} & \textbf{Dev} & \textbf{Test}\\
\hline
WSJ          &POS Tagging   &912344  & 131768 & 129654 \\
CoNLL 2000    &Chunking       &211727  & -      & 47377 \\
CoNLL 2003    &NER           &204567  & 51578  & 46666 \\
\hline

\end{tabular}
\caption{Statistics of four multi-task datasets for sequence tagging.}
\label{tab:seq_data}
\end{table}

\begin{table}[t]\setlength{\tabcolsep}{2pt} \small
\centering
\begin{tabular}{l*{3}{c}}
\hline
 & {\footnotesize CoNLL2000$^\dag$} &{\footnotesize 
CoNLL2003$^\dag$} &{\footnotesize WSJ$^\ddag$} \\\hline
Single Task Model:\\
LSTM+CRF$^\blacklozenge$  & 93.67 &89.91 &97.25 \\
Meta-LSTM+CRF &93.71 &\textbf{90.08}  &\textbf{97.30}\\
\citet{collobert2011natural}     &94.32  &89.59 &97.29 \\
\hline
Multi-Task Model:\\
LSTM-SSP-MTL+CRF & 94.32 &90.38 &97.23\\
Meta-LSTM-MTL+CRF   &\textbf{95.11} &\textbf{90.72} &\textbf{97.45} \\
\bottomrule
\end{tabular}
\caption{
Accuracy rates of our models on three tasks for sequence tagging.$\dag$ means evaluated by F1 score($\%$), $\ddag$ means evaluated by accuracy($\%$). $\blacklozenge$ is the model implemented in \cite{huang2015bidirectional} .
}\label{tab:mtl-seq_tag}
\end{table}

\subsection{Exp-II: Multi-task Learning of  Sequence Tagging}
In this section, we conduct experiment for sequence tagging. Similar to \cite{huang2015bidirectional,ma2016end}, we use the bi-directional Meta-LSTM layers to encode the sequence and a conditional random field (CRF) \cite{lafferty2001conditional} as output layer.
 The hyperparameters settings are same to Exp-I, but with 100d embedding size and 30d Meta-LSTM size.

\paragraph{Datasets}
For sequence tagging task, we use the Wall Street Journal(WSJ) portion of Penn Treebank (PTB) \cite{marcus1993building}, CoNLL 2000 chunking, and CoNLL 2003 English NER datasets. The statistics of these datasets are described in Table \ref{tab:seq_data}.

\paragraph{Experiment result}
Table \ref{tab:mtl-seq_tag} shows the accuracies or F1 scores on the sequence tagging datasets of our models, compared to some state-of-the-art results. As shown, our proposed Meta-LSTM performs better than our competitor models whether it is single or multi-task learning. 

\subsection{Result Analysis}
From the above two experiments, we have empirically observed that our model is consistently better than the competitor models, which shows our model is very robust. Explicit to multi-task learning, our model outperforms SSP-MTL and PSP-MTL by a large margin with fewer parameters, which indicates the effectiveness of our proposed functional sharing mechanism.

\section{Related Work}

One thread of related work is neural networks based multi-task learning, which has been proven effective in many NLP problems \cite{collobert2008unified,glorot2011domain,liu2015representation,liu2016recurrent}. In most of these models, the lower layers are shared across all tasks, while top layers are task-specific.
This kind of sharing scheme divide the feature space into two parts: the shared part and the private part. The shared information is representation-level, whose capacity grows linearly as the size of shared layers increases.

Different from these models, our model captures the function-level sharing information, in which a meta-network captures the meta-knowledge across tasks and controls the parameters of task-specific networks.

Another thread of related work is the idea of using one network to predict the parameters of another network. \citet{de2016dynamic} used a filter-generating network to generate the parameters of another dynamic filter network, which implicitly learn a variety of filtering operations.
\citet{bertinetto2016learning} introduced a learnet for one-shot learning, which can predicts the parameters of a second network given a single exemplar.
\citet{ha2016hypernetworks} proposed the model hypernetwork, which uses a small network to generate the weights for a larger network. In particular, their proposed hyperLSTMs is same with our Meta-LSTMs except for the computational formulation of the dynamic parameters. Besides, we also use a \textit{low-rank approximation} to generate the parameter matrix, which can reduce greatly the model complexity, while keeping the model ability.

\section{Conclusion and Future Work}

In this paper, we introduce a novel knowledge sharing scheme for multi-task learning. The difference from the previous models is the mechanisms of sharing information among several tasks. We design a meta network to store the knowledge shared by several related tasks. With the help of the meta network, we can obtain better task-specific sentence representation by utilizing the knowledge obtained by other related tasks.
Experimental results show that our model can improve the performances of several related tasks by exploring common features and outperforms the representational sharing scheme.
The knowledge captured by the meta network can be transferred across other new tasks.

In future work, we would like to investigate other functional sharing mechanisms of neural network based multi-task learning.

\section*{Acknowledgement}
We would like to thank the anonymous reviewers for their valuable comments. The research work is supported by the National Key Research and Development Program of China (No. 2017YFB1002104), Shanghai Municipal Science and Technology Commission (No. 17JC1404100), and National Natural Science Foundation of China (No. 61672162).

\bibliographystyle{aaai}
\bibliography{nlp}

\end{document}